\documentclass[sigconf]{aamas} 

\usepackage{balance} 
\usepackage{hyperref}       
\usepackage{booktabs}       
\usepackage{amsfonts,amsmath}       
\usepackage{nicefrac}       
\usepackage{microtype}      
\usepackage{subcaption}
\usepackage[capitalize]{cleveref}
\usepackage{siunitx}

\usepackage[noend]{algpseudocode}
\usepackage{pgfplots, pgfplotstable}
\usetikzlibrary{pgfplots.groupplots}
\pgfplotsset{compat=1.16}
\usetikzlibrary{shapes.arrows, fadings, shapes}
\usetikzlibrary{arrows}
\usetikzlibrary{positioning}
\definecolor{nice-red}{HTML}{E41A1C}
\definecolor{nice-orange}{HTML}{FF7F00}
\definecolor{nice-yellow}{HTML}{FFC020}
\definecolor{nice-green}{HTML}{4DAF4A}
\definecolor{nice-blue}{HTML}{377EB8}
\definecolor{nice-purple}{HTML}{984EA3}
\definecolor{nice-grey}{HTML}{6C7A89}
\definecolor{nice-pink}{HTML}{DB5A6B}

\definecolor{color0}{rgb}{0.00392156862745098, 0.45098039215686275, 0.6980392156862745}
\definecolor{color1}{rgb}{0.8705882352941177, 0.5607843137254902, 0.0196078431372549}
\definecolor{color2}{rgb}{0.00784313725490196, 0.6196078431372549, 0.45098039215686275}
\definecolor{color3}{rgb}{0.8352941176470589, 0.3686274509803922, 0.0}
\definecolor{color4}{rgb}{0.8, 0.47058823529411764, 0.7372549019607844}
\definecolor{color5}{rgb}{0.792156862745098, 0.5686274509803921, 0.3803921568627451}
\definecolor{color6}{rgb}{0.984313725490196, 0.6862745098039216, 0.8941176470588236}
\definecolor{color7}{rgb}{0.5803921568627451, 0.5803921568627451, 0.5803921568627451}
\definecolor{color8}{rgb}{0.9254901960784314, 0.8823529411764706, 0.2}
\definecolor{color9}{rgb}{0.33725490196078434, 0.7058823529411765, 0.9137254901960784}
\usepackage{tikz}
\usepackage{doi}
\usepackage{wrapfig}
\usepackage{footnote}
\makesavenoteenv{tabular}
\usepackage{tablefootnote}
\usepackage{threeparttable}

\setcopyright{ifaamas}
\acmConference[AAMAS '21]{Proc.\@ of the 20th International Conference on Autonomous Agents and Multiagent Systems (AAMAS 2021)}{May 3--7, 2021}{London, UK}{U.~Endriss, A.~Now\'{e}, F.~Dignum, A.~Lomuscio (eds.)}
\copyrightyear{2021}
\acmYear{2021}
\acmDOI{}
\acmPrice{}
\acmISBN{}

\acmSubmissionID{318}

\title[AAMAS-2021 Formatting Instructions]{Deep Implicit Coordination Graphs for \\ Multi-agent Reinforcement Learning}

\author{Sheng Li}
\affiliation{
  \institution{Stanford University}
  \city{Stanford}
  \state{CA}}
\email{lisheng@stanford.edu}

\author{Jayesh K. Gupta}
\affiliation{
  \institution{Stanford University}
  \city{Stanford}
  \state{CA}}
\email{jkg@cs.stanford.edu}

\author{Peter Morales}
\affiliation{
  \institution{Microsoft}
  \city{Redmond}
  \state{WA}}
\email{pmorales@microsoft.com}

\author{Ross Allen}
\affiliation{
  \institution{MIT Lincoln Labs}
  \city{Lexington}
  \state{MA}}
\email{ross.allen@ll.mit.edu}

\author{Mykel J. Kochenderfer}
\affiliation{
  \institution{Stanford University}
  \city{Stanford}
  \state{CA}}
\email{mykel@stanford.edu}

\begin{abstract}
Multi-agent reinforcement learning (MARL) requires coordination to efficiently solve certain tasks. 
Fully centralized control is often infeasible in such domains due to the size of joint action spaces.
Coordination graph formalizations allow reasoning about the joint action based on the structure of interactions. 
However, they often require domain expertise in their design and can be difficult for dynamic environments with changing coordination requirements. 
This paper introduces the \emph{deep implicit coordination graph} (DICG) architecture for such scenarios.
DICG consists of a module for inferring the dynamic coordination graph structure which is then used by a graph neural network module to learn to implicitly reason about the joint actions or values.
DICG allows learning the tradeoff between full centralization and decentralization via standard actor-critic methods to significantly improve coordination for domains with large number of agents.
We apply DICG to both centralized-training-centralized-execution and centralized-training-decentralized-execution regimes.
We demonstrate that DICG solves the \emph{relative overgeneralization} pathology in predatory-prey tasks as well as outperforms various MARL baselines on the challenging StarCraft II Multi-agent Challenge (SMAC) and traffic junction environments.
\end{abstract}


\newcommand{\BibTeX}{\rm B\kern-.05em{\sc i\kern-.025em b}\kern-.08em\TeX}

\begin{document}


\pagestyle{fancy}
\fancyhead{}


\maketitle 

\section{Introduction}
\label{sec:intro}
Effective multi-agent reinforcement learning (MARL) in fully cooperative environments often requires coordination between agents on a team.
One simple approach for achieving coordination is to reduce the problem to a single agent problem where the action space is the joint action space of all agents.
Unfortunately, this joint action space grows exponentially with the number of agents, making it intractable for many domains of interest.
To avoid this problem, a common strategy is to \emph{decentralize} or factorize the decision policy or value function for each agent~\cite{tan1993multi,gupta2017cooperative,sukhbaatar2016learning}. 
Each agent selects actions to maximize its corresponding utility function, with the end goal of maximizing the joint value function.
However, such decentralization can be suboptimal \citep{matignon2012}.
The optimal policy is often not learnable in such a context due to a game-theoretic pathology referred to as \emph{relative overgeneralization} \citep{panait2006biasing}, where the agent's reward gets confounded by penalties from random exploratory actions of other collaborating agents.

\citet{guestrin2002multiagent} introduced the framework of \emph{coordination graph} (CG) to reason about joint value estimates from a factored representation to significantly improve computational tractability at the expense of optimality.
Compared to function decomposition schemes like Value Decomposition Networks (VDN)~\citep{sunehag2018value}, QMIX~\citep{rashid2018qmix}, and parameter sharing in decentralized policy optimization~\citep{gupta2017cooperative}, the CG framework allows explicit modeling of the locality of interactions and formal reasoning about joint actions given the coordination graph structure.
\citet{kok2004sparse} applied these ideas in the context of tabular reinforcement learning. The approach was later extended to function approximation with neural networks by \citet{Bohmer2019-zv}.
Most of these approaches assume a domain dependent static coordination graph is given. 
Although the coordination graph terminology is focused on using a graph data-structure to decompose payoffs and utilities, we believe the idea of using a graph data-structure for coordinated action inference can be considered more general. 

For a wide range of problems, this coordination graph structure is dynamic and state dependent.
Domain heuristics like adding a graph edge with neighboring agents based on some distance metric are sometimes used~\citep{Jiang2020Graph}.
The approach of \citet{kok2005utile} attempts to learn such structure, but it is limited to tabular settings with domain heuristics.
We hypothesize that methods that learn the appropriate dynamic coordination graph to inform the selection of joint actions can help address coordination issues in MARL.

We propose the \emph{Deep Implicit Coordination Graph} (DICG) module for multi-agent deep RL. 
It uses a self-attention network to determine, what we call, an implicit coordination graph structure which is then used for agent information integration through a graph convolutional network (GCN). 
Although ``implicit coordination graph'' is not strictly the payoff-utility based coordination graph as is standard in the literature, it builds off the same idea of reasoning about the joint action based on the relatively sparse interactions between the agents.
The intuition behind this architecture design is to make both the coordination graph structure and action inference over its edges differentiable so that the entire DICG module can be used inside either the actor or the critic and trained end-to-end through standard policy optimization methods. 
Since the module is trained to optimize the joint reward, the GCN submodule learns to implicitly reason about joint actions/values based on the structure of interaction inferred by the attention submodule.

We compare DICG with fully centralized and decentralized MARL methods in a challenging domain involving predator-prey tasks that require strong coordination. 
We also study performance on the StarCraft II Multi-Agent Challenge (SMAC)~\citep{samvelyan2019starcraft} and the traffic junction environment~\citep{sukhbaatar2016learning}.
DICG learns the relevant dynamic coordination graph structure, allowing it to make an appropriate trade-off between centralized and decentralized methods. 
\section{Background and Related Work}
\label{sec:bg}
We formalize the problem as a Dec-POMDP \citep{oliehoek2016concise} forming the tuple $\langle \mathcal{I}, \mathcal{S}, \{\mathcal{A}^i\}_{i=1}^n, \mathcal{T}, \mathcal{Z}, R, \mathcal{O}, \gamma\rangle$, where $\mathcal{I} = \{1, \ldots, n\}$ is the set of agents, $\mathcal{S}$ is the global state space, $\mathcal{A}^i$ is the action space of the $i$th agent, and $\mathcal{Z}$ is the observation space for an agent.
The transition function defining the next state distribution is given by $\mathcal{T}: \mathcal{S} \times \prod_i \mathcal{A}^i \times \mathcal{S} \to [0,1]$.
The reward function is $R: \mathcal{S} \times \prod_i \mathcal{A}^i \to \mathbb{R}$, and the discount factor is $\gamma \in [0, 1)$.
The observation model defining the observation distribution from the current state is $\mathcal{O}: \mathcal{S} \times \mathcal{Z} \to [0, 1]$.
Each agent $i$ has a stochastic policy $\pi^i$ conditioned on its observations $o_i$ or action-observation history $\tau^i \in (\mathcal{Z} \times \mathcal{A}^i)$.
The discounted return is $G_t = \sum_{l=0}^\infty \gamma^{l} r_{t+l}$, where $r_t$ is the joint reward at step $t$.
The joint policy $\mathbf{\pi}$ induces a value function $V^{\mathbf{\pi}}(s_t) = \mathbb{E}[G_t \mid s_t]$ and an action-value function $Q^\mathbf{\pi}(s_t, \mathbf{a}_t) = \mathbb{E}[G_t \mid s_t, \mathbf{a}_t]$, where $\mathbf{a}_t$ is the joint action.
The advantage function is then $A^\mathbf{\pi}(s_t, \mathbf{a}_t)=Q^\mathbf{\pi}(s_t, \mathbf{a}_t) - V^{\mathbf{\pi}}(s_t)$.

\subsection{Policy Optimization}
We use policy optimization to maximize the expected discounted return.
Given policy $\pi_\theta$ parameterized by $\theta$, the surrogate policy optimization objective is~\citep{schulman2017proximal}:
\begin{equation}
    \underset{\theta}{\mathrm{maximize}} \quad \hat{\mathbb{E}}_t\left[\frac{\pi_{\theta}(a_t \mid s_t)}{\pi_{\theta_{\text{old}}}(a_t \mid s_t)} \hat{A}_t\right]
\end{equation} 
where $\hat{A}_t$ is the advantage function estimator~\citep{schulman2015high} at time step $t$ and the expectation $\hat{\mathbb{E}}_t[\ldots]$ indicates the empirical average over a finite batch of samples.
In practice, we use the clipped PPO objective \citep{schulman2017proximal} to limit the step size for stable updates.
For the policy $\pi_\theta$, we can either condition on states using a feed-forward network like multi-layer perceptron (MLP), or condition on the full history using a recurrent neural network such as an LSTM~\citep{hochreiter1997long} or GRU~\citep{chung2014empirical}. 
In the context of centralized training but decentralized execution, a common strategy is to share the policy parameters between agents that are homogeneous~\citep{gupta2017cooperative,Bohmer2019-zv}.
With shared rewards, COMA~\citep{foerster2018counterfactual} critic can be useful for better credit assignment and can be easily combined with our proposed approach.
However, these approaches do not model the coordination structure.
\citet{wei2018multiagent} investigate relative overgeneralization in continuous action multi-agent tasks and show improvement over MADDPG~\citep{lowe2017multi}.
\citet{oroojlooyjadid2019review} provide a general overview of cooperative MARL.

\subsection{Coordination Graphs}
For several multi-agent domains, the outcome of an agent's action often depends only on a subset of other agents in the domain. 
This \emph{locality of interaction} can be encoded in the form of a coordination graph (CG)~\citep{guestrin2002multiagent}.
A CG is often represented as an undirected graph $G=\langle \mathcal{V}, \mathcal{E}\rangle$ and contains a vertex $v_i \in \mathcal{V}$ for each agent $i$ and a set of undirected edges $\{i,j\} \in \mathcal{E}$ between vertices $v_i$ and $v_j$.
\citet{guestrin2002multiagent} use this CG to induce a factorization of an action-value function into \emph{utility functions} $f^i$ and \emph{payoff functions} $f^{ij}$:
\begin{equation}
    q^{\mathrm{CG}}(s_t, a) = \sum_{v^i \in \mathcal{V}} f^i(a^i \mid s_t) +  \sum_{\{i,j\} \in \mathcal{E}} f^{ij}(a^i, a^j \mid s_t)
\end{equation}
\citet{guestrin2002multiagent} and \citet{vlassis2004anytime} draw on the connections with maximum a posteriori (MAP) estimation techniques in probabilistic inference to compute the joint action from such factorizations; resulting into algorithms like Variable Elimination and Max-Plus. \citet{kok2004sparse} explored their use in the context of tabular MARL. Deep Coordination Graphs (DCG)~\citep{Bohmer2019-zv} extended these ideas of factoring the joint value function of all agents according to a static coordination graph into payoffs between pairs of agents to deep MARL. 
They did so by estimating the payoff functions using neural networks and using message passing based on Max-Plus~\citep{vlassis2004anytime} along the coordination graph to maximize the value function, allowing training of the value function end-to-end.

In this work, however, we forgo explicitly computing the joint action through inference over factored representation with a \emph{given} coordination graph.
Instead, we use attention to learn the appropriate agent observation-dependent coordination graph structure with soft edge weights and then use message passing in a graph neural network to compute appropriate values or actions for the agents, such that full the computation graph remains differentiable.


\subsection{Self-attention}

Self-attention mechanism~\citep{cheng2016long} emerged from the natural language processing community.
It is used to relate different positions of a single sequence. The difference between self-attention and standard attention is that self-attention uses a single sequence as both its source and target sequence. It has been shown useful in image caption generation~\citep{liu2018show, yu2019multimodal} and machine reading~\citep{cheng2016long, yu2018qanet}.

The attention mechanism has also been adopted recently for MARL. The relations between a group of agents can be learned through attention. \citet{iqbal2018actor} use attention to extract relevant information of each agent from the other agents. \citet{jiang2018learning} use self-attention to learn when to communicate with neighboring agents. \citet{wright2019attentional} use self-attention on the policy level to differentiate different types of connections between agents. \citet{Jiang2020Graph} use multi-head dot product attention to compute interactions between neighbouring agents for the purpose of enlarging agents' receptive fields and extracting latent features of observations. 
We use self-attention to learn the attention weights between agents, and use the attention weights to form a ``soft''-edged coordination graph instead of edges with binary weights.

\begin{figure*}[t]
    \centering
    \resizebox{5.8in}{3.83in}{%

\begin{tikzpicture}[>=stealth']
\node[rectangle, draw=nice-green, fill=nice-green!5, minimum height=1.5cm, minimum width=7.5cm] (_encoder) at (0, 0.2) {};
\node[] () at (-0.3, 0.6) {\textbf{Encoder}};

\node[trapezium, draw=nice-green, fill=nice-green!30, align=center, inner xsep=8pt, inner ysep=6pt, shape border rotate=180] (_encoder_n) at (3, 0) {};
\node[trapezium, draw=nice-green, fill=nice-green!30, align=center, inner xsep=8pt, inner ysep=6pt, shape border rotate=180] (_encoder_i) at (0.75, 0) {};
\node[trapezium, draw=nice-green, fill=nice-green!30, align=center, inner xsep=8pt, inner ysep=6pt, shape border rotate=180] (_encoder_2) at (-1.5, 0) {};
\node[trapezium, draw=nice-green, fill=nice-green!30, align=center, inner xsep=8pt, inner ysep=6pt, shape border rotate=180] (_encoder_1) at (-3, 0) {};

\node[align=center] at (-0.375, 0) {$\dots$};
\node[align=center] at (1.875, 0) {$\dots$};


\node[rectangle, draw=black, fill=black!5, minimum height=1cm, minimum width=7cm] (_observations) at (0, -2) {};

\node[circle, draw=black, fill=black!10, align=center] (_o1) at (-3, -2) {$o_1$};
\node[circle, draw=black, fill=black!10, align=center] (_o2) at (-1.5, -2) {$o_2$};
\node[circle, draw=black, fill=black!10, align=center] (_oi) at (0.75, -2) {$o_i$};
\node[circle, draw=black, fill=black!10, align=center] (_on) at (3, -2) {$o_n$};

\node[align=center] at (1.875, -2) {$\dots$};
\node[align=center] at (-0.375, -2) {$\dots$};
\draw[->, line width=0.3mm, draw=nice-grey] (_o1) to (_encoder_1);
\draw[->, line width=0.3mm, draw=nice-grey] (_o2) to (_encoder_2);
\draw[->, line width=0.3mm, draw=nice-grey] (_oi) to (_encoder_i);
\draw[->, line width=0.3mm, draw=nice-grey] (_on) to (_encoder_n);


\node[rectangle, draw=nice-yellow, fill=nice-yellow!5, minimum height=2.4cm, minimum width=7.5cm] (_attention) at (0, 2.5) {};
\node[] () at (-2.6, 3.4) {\textbf{Attention}};

\node[circle, draw=nice-blue, fill=nice-blue!30, align=center] (_e1) at (-3, 2) {$e_1$};
\node[circle, draw=nice-blue, fill=nice-blue!30, align=center] (_e2) at (-1.5, 2) {$e_2$};
\node[circle, draw=nice-blue, fill=nice-blue!30, align=center] (_ei) at (0.75, 2) {$e_i$};
\node[circle, draw=nice-blue, fill=nice-blue!30, align=center] (_en) at (3, 2) {$e_n$};

\node[align=center] at (1.875, 2) {$\dots$};
\node[align=center] at (-0.375, 2) {$\dots$};

\draw[->, line width=0.3mm, draw=nice-grey] (_encoder_1) to (_e1);
\draw[->, line width=0.3mm, draw=nice-grey] (_encoder_2) to (_e2);
\draw[->, line width=0.3mm, draw=nice-grey] (_encoder_i) to (_ei);
\draw[->, line width=0.3mm, draw=nice-grey] (_encoder_n) to (_en);

\draw[->, line width=0.3mm, bend right=60] (_ei) edge [above] node [align=center] {$\mu_{i1}$} (_e1);
\draw[->, line width=0.3mm, bend right=60] (_ei) edge [below] node [align=center] {$\mu_{i2}$} (_e2);
\draw[->, line width=0.3mm, bend left=60] (_ei) edge [above] node [align=center] {$\mu_{in}$} (_en);
\draw[->, line width=0.3mm, loop above]  (_ei) edge [loop above] node [align=center] {$\mu_{ii}$} (_ei);

\node[] (_E) at (1.5, 4) {$M=\{\mu_{ij}\}$, $E^{(0)}=\{e_i\}$};

\node[rectangle, draw=nice-blue, fill=nice-blue!5, minimum height=3.5cm, minimum width=17cm] (_attention) at (4, 7) {};
\node[] () at (4, 5.8) {\textbf{Graph Convolution}};

\node [
    draw=nice-orange,
    bottom color=nice-orange!10,
    top color=nice-orange!70,
    single arrow,
    minimum height=1.7cm,
    minimum width=1.5cm,
    single arrow head extend=0.4cm,
    shape border rotate=90
] at (-1.5, 4.5) {$E^{(0)}$, $M$};

\node[circle, draw=nice-blue, fill=nice-blue!30, align=center] (_e1_gcn) at (-1.5-1.5, 6) {$e_1^{(0)}$};
\node[circle, draw=nice-blue, fill=nice-blue!30, align=center] (_e2_gcn) at (-1.8-1.5, 8) {$e_2^{(0)}$};
\node[circle, draw=nice-blue, fill=nice-blue!30, align=center] (_ei_gcn) at (1.2-1.5, 8) {$e_i^{(0)}$};
\node[circle, draw=nice-blue, fill=nice-blue!30, align=center] (_en_gcn) at (1.5-1.5, 6) {$e_n^{(0)}$};

\draw[<->,line width=0.3mm] (_e1_gcn) to (_e2_gcn);
\draw[<->, line width=0.1mm] (_e1_gcn) to (_en_gcn);
\draw[<->, line width=0.2mm] (_e1_gcn) to (_ei_gcn);
\draw[<->, line width=0.4mm] (_e2_gcn) to (_en_gcn);
\draw[<->, line width=0.5mm] (_e2_gcn) edge [above] node [align=center] {$\mu_{i2}$, $\mu_{2i}$} (_ei_gcn);
\draw[<->, line width=0.6mm] (_ei_gcn) to (_en_gcn);

\node [
    draw=nice-red,
    fill=nice-red!20,
    single arrow,
    minimum height=1.5cm,
    minimum width=1.2cm,
    single arrow head extend=0.4cm,
] at (1.5, 7) {$GCN_1(E^{(0)}, M)$};

\node[] at (3.85, 7) {$E^{(1)}\dots$};

\node [
    draw=nice-red,
    fill=nice-red!20,
    single arrow,
    minimum height=1.5cm,
    minimum width=1.2cm,
    single arrow head extend=0.4cm,
] at (6, 7) {$GCN_m(E^{(m-1)}, M)$};

\node[circle, draw=nice-blue, fill=nice-blue!30, align=center] (_e1_gcn_m) at (-1.5+10, 6) {$e^{(m)}_1$};
\node[circle, draw=nice-blue, fill=nice-blue!30, align=center] (_e2_gcn_m) at (-1.8+10, 8) {$e^{(m)}_2$};
\node[circle, draw=nice-blue, fill=nice-blue!30, align=center] (_ei_gcn_m) at (1.2+10, 8) {$e^{(m)}_i$};
\node[circle, draw=nice-blue, fill=nice-blue!30, align=center] (_en_gcn_m) at (1.5+10, 6) {$e^{(m)}_n$};

\draw[<->,line width=0.3mm] (_e1_gcn_m) to (_e2_gcn_m);
\draw[<->, line width=0.1mm] (_e1_gcn_m) to (_en_gcn_m);
\draw[<->, line width=0.2mm] (_e1_gcn_m) to (_ei_gcn_m);
\draw[<->, line width=0.4mm] (_e2_gcn_m) to (_en_gcn_m);
\draw[<->, line width=0.5mm] (_e2_gcn_m) edge [above] node [align=center] {$\alpha_{i2}$, $\alpha_{2i}$} (_ei_gcn_m);
\draw[<->, line width=0.6mm] (_ei_gcn_m) to (_en_gcn_m);

\node [
    draw=nice-orange,
    bottom color=nice-orange!10,
    top color=nice-orange!70,
    single arrow,
    minimum height=1.1cm,
    minimum width=1.1cm,
    single arrow head extend=0.2cm,
    shape border rotate=270
] at (10, 5) {};

\node[] (_E_m) at (10, 4) {$E^{(m)}=\{e^{(m)}_i\}$};

\node[circle, draw=black] (_+) at (6.5, 4) {+};
\node[] (_E_til) at (6.5, 2.8) {$\tilde{E}=\{\tilde{e}_i\}$};

\draw[->,line width=0.3mm, draw=nice-grey] (_E) to (_+);
\draw[->,line width=0.3mm, draw=nice-grey] (_E_m) to (_+);
\draw[->,line width=0.3mm, draw=nice-grey] (_+) to (_E_til);

\node[rectangle, draw=nice-purple, fill=nice-purple!5, minimum height=1cm, minimum width=3cm, text=blue] () at (8.3, 1.3) {};
\node[rectangle, draw=nice-purple, fill=nice-purple!5, minimum height=1cm, minimum width=3cm, text=blue] (_policy_ctce) at (8.2, 1.2) {};
\node[rectangle, draw=nice-purple, fill=nice-purple!5, minimum height=1cm, minimum width=3cm, text=blue] () at (8.1, 1.1) {};
\node[rectangle, draw=nice-purple, fill=nice-purple!10, minimum height=1cm, minimum width=3cm, text=nice-blue] () at (8, 1) {{Policy (CTCE)}};

\node[rotate=45] () at (8.65, 2.05) {$\dots$};

\draw[->, line width=0.3mm, draw=nice-blue, rounded corners=5pt] (_E_til) -| (_policy_ctce);

\node[rectangle, draw=nice-pink, fill=nice-pink!5, minimum height=1cm, minimum width=3cm, text=blue] () at (8.1, -1.9) {};
\node[rectangle, draw=nice-pink, fill=nice-pink!5, minimum height=1cm, minimum width=3cm, text=blue] (_policy_ctde) at (8, -2) {};
\node[rectangle, draw=nice-pink, fill=nice-pink!5, minimum height=1cm, minimum width=3cm, text=blue] () at (7.9, -2.1) {};
\node[rectangle, draw=nice-pink, fill=nice-pink!10, minimum height=1cm, minimum width=3cm, text=nice-red] () at (7.8, -2.2) {{Policy (CTDE)}};

\node[rotate=45] () at (8.45, -1.15) {$\dots$};

\node[rectangle, draw=cyan, fill=cyan!20, minimum height=1cm, minimum width=1cm] (_aggregator) at (5, -0.5) {Aggregator};

\draw[->, draw=nice-red, rounded corners=5pt, line width=0.3mm] (_E_til) -| (_aggregator);
\draw[->, draw=nice-red, rounded corners=5pt, line width=0.3mm] (_aggregator) -| (_policy_ctde);

\draw[->, draw=nice-red, line width=0.3mm] (_observations) edge [above] node [align=center] {Observations} (6.2, -2);

\node[] () at (7, -0.2) {Baseline};

\node[] (_actions_ctde) at (11, -2) {Actions};
\node[] (_actions_ctce) at (11, 1.2) {Actions};

\draw[->,draw=nice-red,line width=0.3mm] (_policy_ctde) to (_actions_ctde);
\draw[->,draw=nice-blue,line width=0.3mm] (_policy_ctce) to (_actions_ctce);

\end{tikzpicture}

}
    \caption{Network architecture of DICG. It can be used for either a centralized-training-centralized-execution (CTCE) approach or as a centralized-training-decentralized-execution (CTDE) approach. The \textcolor{nice-blue}{blue} arrows indicate the CTCE approach. The DICG module serves as a joint observation encoder. We use the integrated observations $\Tilde{E}$ to directly obtain actions for agents through a parameter sharing policy. The baselines in CTCE are estimated by a concatenation of raw observations. The \textcolor{nice-red}{red} arrows indicate the CTDE approach. We pass the integrated observations $\Tilde{E}$ through an aggregator network to estimate a centralized baseline. We then use the baseline to compute the advantage to guide policy optimization.}
    \label{fig:arch}
\end{figure*}
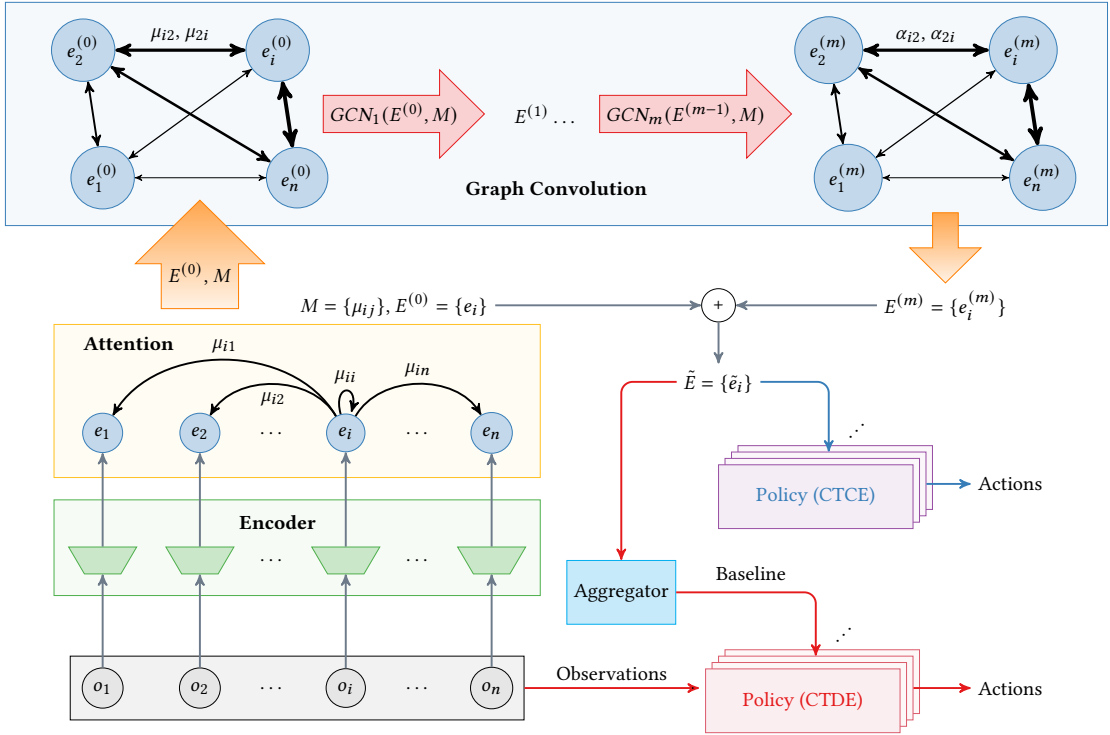

\subsection{Graph Neural Networks}
Several frameworks have been proposed to extract locally connected features from arbitrary graphs~\citep{kipf2016semi,velickovicgraph}.
Given a graph $G=\langle \mathcal{V}, \mathcal{E}\rangle$, a graph convolutional network (GCN) takes as input the feature matrix that summarizes the attributes of each node $v_i \in \mathcal{V}$ and outputs a node-level feature matrix. 
This is similar to how a convolution operation across local regions of the input produces feature maps in CNNs. 
MAGnet learns relation weights through a loss function based on heuristic rules in the form of a relevance graph~\citep{malysheva2018deep}.
Deep relational RL embeds multi-head dot-product attention as relational block into graph neural networks to learn pairwise interaction representation of a set of entities in the agent's state~\citep{zambaldi2018relational}.
Recently, \citet{liu2019multi} combined a two-stage attention network with a graph neural network for communication between the agents to achieve state-of-the-art performance on the traffic junction domain with curriculum training~\citep{bengio2009curriculum}.
\section{Approach} \label{sec:approach}
Instead of the standard approach of learning the binary weights of the edges in a coordination graph, we use self-attention to learn the relation between agents and use the attention weights as soft edges of a coordination graph. These soft edges form an implicit coordination graph representing elements of its adjacency matrix, $M \in \mathbb{R}_{>0}^{n \times n}$. 
We use self-attention to avoid building coordination graphs using hard-coded or domain-specific heuristics 
so that our approach is applicable to more abstract multi-agent domains. 
Moreover, maintaining differentiability is difficult with binary connections. We use attention to implicitly represent the edge weights as the strength of the connection between agents to obtain the graphs's adjacency matrix.
We then apply graph convolution~\citep{kipf2016semi} with this adjacency matrix to integrate information across agents. 
We use graph convolution because it is an efficient and differentiable way to pass information along the graph.
With the integrated information, we can either use it as observation embeddings to directly obtain actions or use it to estimate baselines for advantage estimation during policy optimization.
In summary, the DICG module consists of an encoder, an attention module, and a graph convolution module with the architecture outlined in~\cref{fig:arch}.

In detail, we first pass $n$ observations $\{o_i\}_{i=1}^n$ of the $n$ agents through a parameter sharing encoder parameterized by $\theta_e$.
The encoder outputs $n$ embedding vectors $\{e_i\}_{i=1}^n$, each with size $d$:
\begin{equation}
    e_i = \text{Encoder}(o_i; \theta_e)\text{, for $i = 1,\dots,n$}.
\end{equation}
We then compute the attention weights from agent $i$ to $j$ using these embeddings as:
\begin{equation}
    \mu_{ij} = \frac{\exp(\text{Attention}(e_i, e_j, W_a))}{\sum_{k=1}^{n}\exp(\text{Attention}(e_i, e_k, W_a))}.
\end{equation}
where the attention module is parameterized by $W_a$, which is a trainable $d \times d$ weight matrix. The attention score function we adopt is general attention~\citep{luong2015effective}:
\begin{equation}
    \text{Attention}(e_i, e_j, W_a) = e_j^\top W_a e_i.
\end{equation}
The attention module is also parameter shared among agents. 
We use these attention weights to form an $n\times n$ positive real valued adjacency matrix $M$ with $M_{ij} = \mu_{ij}$, which encodes the implicit coordination graph. 
Since we apply soft-max for attention weights, we have 
$\sum_{j=1}^n \mu_{ij} = 1$.

We stack the embeddings to form an $n\times d$ feature matrix $E$ with the $i$th row being the embedding $e_i^\top$. We denote the $E$ before any graph convolution operations as $E^{(0)}$. With the soft adjacency matrix $M$ and the feature matrix $E^{(0)}$, we can apply graph convolution to perform message passing and information integration across all agents. In the fast approximate GCN by \citet{kipf2016semi}, a graph convolution layer is 
\begin{equation}
    H^{(l+1)} = \sigma\left( \tilde{D}^{-\frac{1}{2}} \tilde{M} \tilde{D}^{-\frac{1}{2}} H^{(l)} W^{(l)}_c\right),
\end{equation}
where $H^{(l)}$ is the feature matrix of convolution layer $l$. In our case, $H^{(0)} = E^{(0)} = [e_1^\top; e_2^\top; \dots; e_n^\top]$.
Diagonal entries of $M$ are already positive from the self-attention weights.
Therefore, unlike \citet{kipf2016semi}, we do not need to add an identity matrix for non-zero self-connections and can set $\tilde{M} = M$.
By their definition, $\tilde{D}_{ii} = \sum_{j=1}^n \tilde{M}_{ij} = \sum_{j=1}^n \mu_{ij} = 1$, i.e. $\tilde{D}$ is simplified to an identity matrix, $I_n$. 
The $d\times d$ matrix $W^{(l)}_c$ is a trainable weight matrix associated with layer $l$, and $\sigma$ is a non-linear activation. 

Replacing $\tilde{M}$ with attention weights $M$ and $\tilde{D}$ with $I_n$, the graph convolution operation simplifies to
\begin{equation}
    H^{(l+1)} = \sigma\left( M H^{(l)} W^{(l)}_c \right).
\end{equation}
This graph convolution operation is performed $m$ times. 
We denote the output of $m$th layer $H^{(m)}$ as $E^{(m)}$, which is a stack of integrated embeddings.

\begin{figure*}[t]
     \centering
     \begin{subfigure}[b]{0.3\textwidth}
         \centering
         \includegraphics[width=0.6\textwidth]{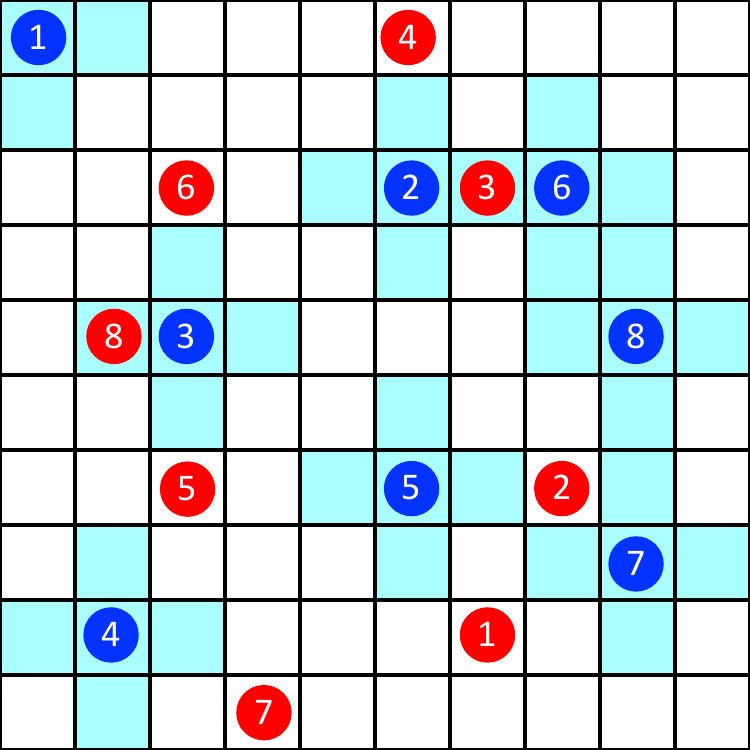}
         \caption{Predator-Prey}
         \label{fig:predprey_env}
     \end{subfigure}
     \begin{subfigure}[b]{0.3\textwidth}
         \centering
         \includegraphics[width=0.6\textwidth]{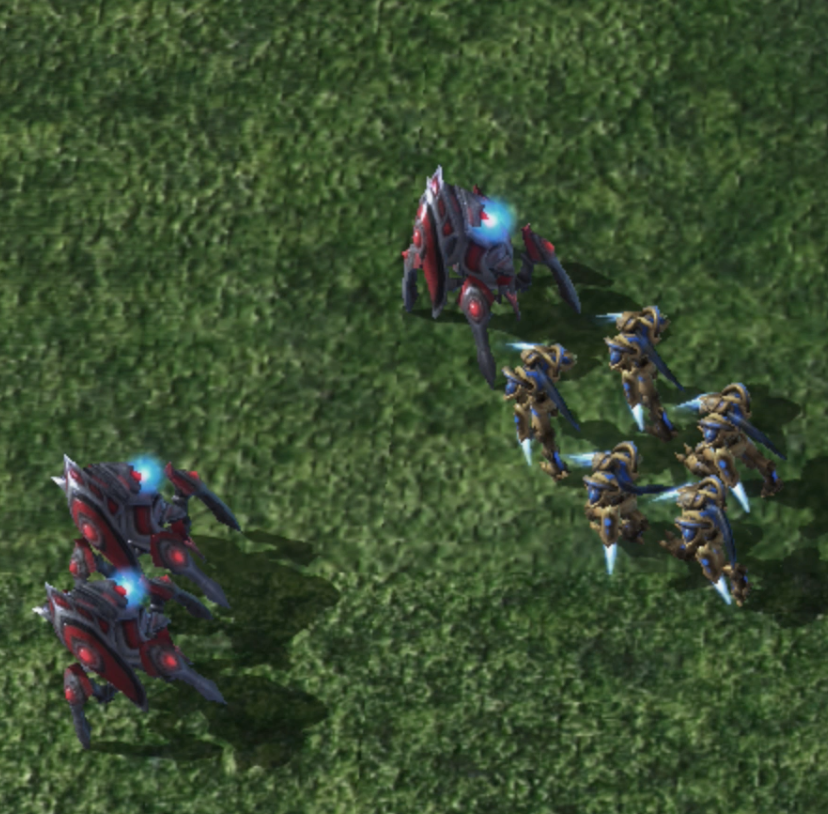}
         \caption{SMAC (\texttt{3s\_vs\_5z})~\cite{vinyals2017starcraft, vinyals2019grandmaster}}
         \label{fig:smac_env}
     \end{subfigure}
     \begin{subfigure}[b]{0.3\textwidth}
         \centering
         \includegraphics[width=0.6\textwidth]{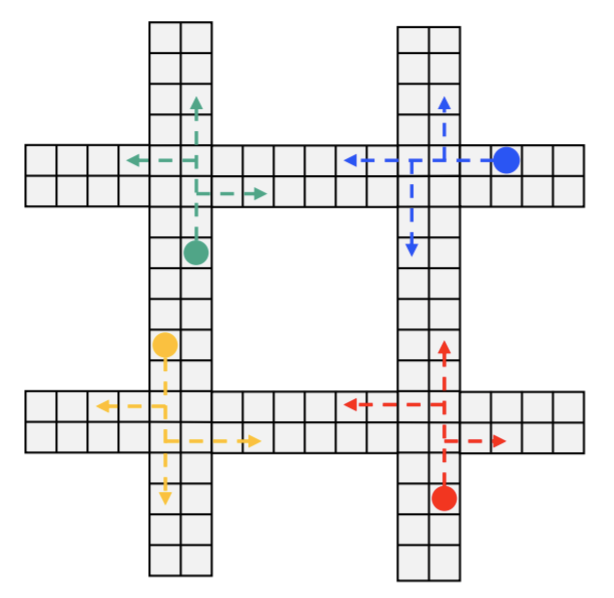}
         \caption{Traffic Junction (hard mode)~\cite{sukhbaatar2016learning, liu2019multi}}
         \label{fig:traffic_env}
     \end{subfigure}
    
    \caption{Experiment environments. (\textbf{a}) Predators are marked in blue, and prey are marked in red. The cyan grids are the capture range of predators. An example of successful capture is predator 2 and 6 capturing prey 3. An example of a single-agent capture attempt that will cause penalty is predator 3 capturing prey 8 alone. (\textbf{b}) SMAC scenario \texttt{3s\_vs\_5z}. (\textbf{c}) An illustration of the hard mode traffic junction environment with $18\times18$ grids and 4 two-way routes.}
    \label{fig:envs}
\end{figure*}

We then use a residual connection~\citep{he2016deep} between $E^{(0)}$ and $E^{(m)}$ to obtain the final embedding matrix $\tilde{E} = E^{(0)} + E^{(m)}$. The residual connection is designed to assist gradient flow through the attention module and the encoder. The final embedding matrix $\tilde{E}$ consists of a stack of integrated embeddings $\{\tilde{e}_i\}^n_{i=1}$. Finally, there are two ways to use the embedding matrix $\tilde{E}$:

    \textbf{(a) DICG-CE}: If full communication is allowed between agents, we can use the DICG module in a \emph{centralized-training-centralized-execution} (CTCE) framework to communicate information between the agents.
    The output from the DICG module, $\tilde{E}$, integrates relevant information across all agents. The corresponding embedding $\tilde{e}_i$ (or its history for recurrent neural networks) can be passed through a separate or parameter-shared policy network to obtain actions for each agent (indicated with \textcolor{nice-blue}{blue} arrows in~\cref{fig:arch}).
    We can then use standard actor-critic methods to train the network end-to-end.
    As our experiments demonstrate, embeddings obtained from DICG are superior to simply concatenating the raw observations and passing through an MLP network due to the implicit coordination structure reasoning for information integration.
    
    \textbf{(b) DICG-DE}: If full communication is not allowed between agents, we can still use the DICG module to facilitate better coordination. Following the principles of \emph{centralized-training-decentralized-execution} (CTDE), we can use the output of the DICG module, $\tilde{E}$, in a centralized critic.
    We pass $\tilde{E}$ through another MLP, which we refer to as an aggregator network, to estimate the centralized baseline (indicated with \textcolor{nice-red}{red} arrows in~\cref{fig:arch}).
    Separate or parameter shared policy networks can be trained for each agent using standard actor-critic methods \citep{gupta2017cooperative,foerster2018counterfactual,allen2019health}, except using the DICG centralized baseline for advantage computation. 
    During execution the critic is no longer required and the agents can act independently.
    Again, we find that the embeddings obtained by DICG are superior to simply concatenating the raw observations and passing through an MLP network due to its implicit reasoning about the dynamic coordination structure.

\textbf{Key Differences from Related Approaches}:
Each component module like self-attention and graph convolutions that are key to our approach have been previously used in the literature.
\citet{iqbal2018actor} use attention mechanism in their critic to dynamically attend to other agents. However, they do not use the concept of a coordination graph and process those embeddings with a graph neural network. Moreover, they experiment with fairly small and simple particle environments.
\citet{wright2019attentional} use attention mechanism in their actor to aggregate information from other agents sharing some similarities with DICG-CE. They do not use attention to create a coordination graph structure that can be used by a graph neural network to process the observation embeddings. Moreover, they are focused on a simple simulation of merging vehicles.
\citet{Jiang2020Graph} use attention mechanism to obtain a relational kernel for use in a graph neural network. Moreover they are focused on value based methods and use Deep Q-learning. They do not use the attention weights to create a coordination graph. Rather, they hand craft the adjacency matrix used by the graph neural network. They experiment with particle environments with simple observations.
\citet{ryu2020multi} use both an attention mechanism and graph neural networks. However, they focus on multi-group settings and consider the relationships between an individual agent with groups of other agents to come up with a hierarchical representation. Again, they only experiment with simple particle environments.

The key contribution of this work is to take inspiration from the coordination graph literature and combine these various components in a specific way to design a fully differentiable architecture.
As we'll see in the next section, this design leads to strong performance improvements in a variety of complex multi-agent domains.

\begin{figure*}[t!]
    \centering
    \input{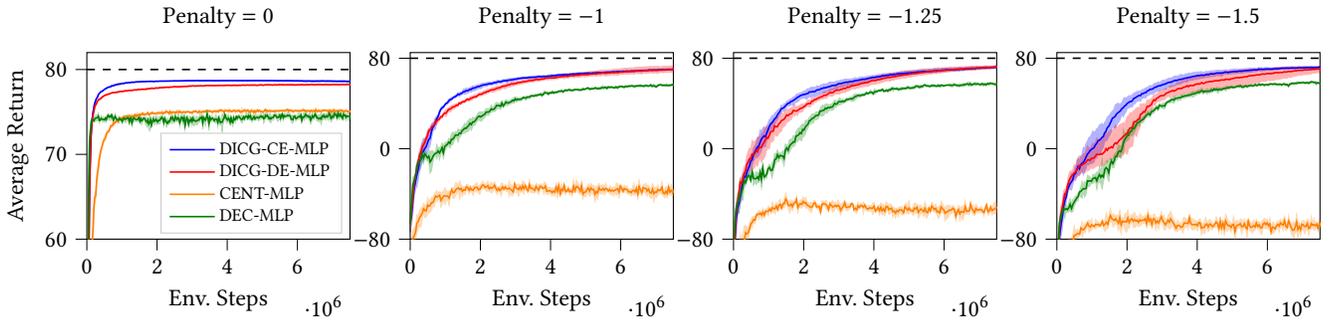}
    \caption{Average return of predator-prey with increasing penalty for single-agent capture attempt.}
    \label{fig:predprey_avgreturn}
\end{figure*}

\section{Results} \label{sec:results}
We present experiments applying DICG to three environments (shown in \cref{fig:envs}): predator-prey, StarCraft II Multi-agent Challenge (SMAC)~\citep{samvelyan2019starcraft}, and traffic junction~\citep{sukhbaatar2016learning}. These environments require coordination to achieve high returns, i.e. agent interactions are not so sparse that totally decentralized approaches with partial observability can achieve high returns. They are also sufficiently complex that fully centralized approaches quickly become intractable. We compare our approach against two standard actor-critic baseline approaches: 1) fully decentralized architecture (referred to as DEC) with only local observation as input to policy; and 2) centralized architecture (referred to as CENT) with a direct concatenation of all observations as input to policy.
Due to the dimensionality of the action space, we still need to factorize the policy~\citep{gupta2017cooperative,sukhbaatar2016learning} in the centralized architecture so that we output separate action distributions for each agent.
They both use a centralized critic with full observation to create a baseline estimate, and they use PPO for policy optimization~\citep{allen2019health}. Benchmarking against them can justify the effectiveness of coordination learning and information integration of DICG. We also compare with results reported by other MARL approaches. 
All results are averaged over $5$ seeds. 
Environment and network details are in Appendix A. Code is available here\footnote{Code of this work: \url{https://github.com/sisl/DICG}}.

\begin{figure*}[t]
    \centering
    \input{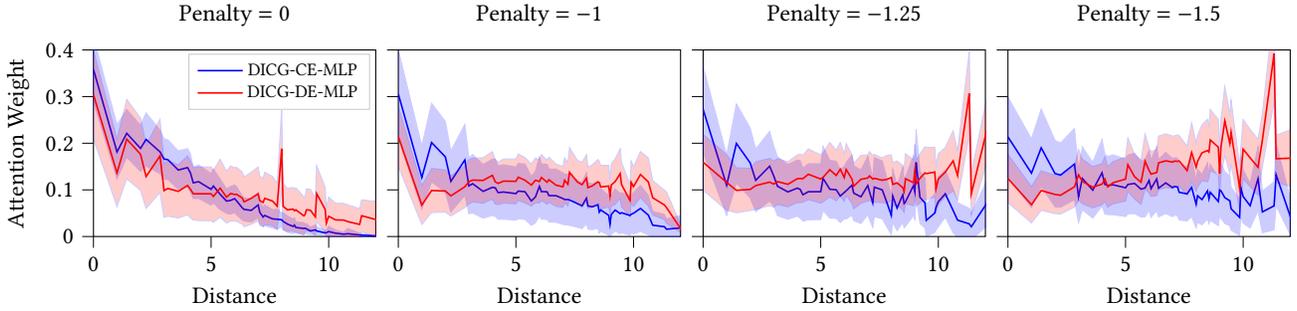}
    \caption{Attention weights, i.e. graph edge strengths of DICG under different distance between two agents (average of 5 seeds). Zero distance indicates an agent's attention towards itself. As the penalty for single-agent capture attempts increases, more attention is paid to farther agents.}
    \label{fig:predprey_attnweights}
\end{figure*}

\subsection{Predator-Prey}
We use an environment similar to that described by \citet{Bohmer2019-zv}. The environment consists of a $10\times10$ grid world with 8 predators and 8 prey. We control the movement of predators to capture prey. The prey move by hard-coded and randomized rules to avoid predators. If a prey is captured, the agents receives a reward of $10$. However, the environment penalizes any single-agent attempt to capture prey with a negative reward $p$; at least two agents are required to be present in the neighboring grid cells of a prey for a successful capture. We set the episode length to $200$ steps, and impose a step cost of $-0.1$. Cooperation is necessary to achieve a high return in this environment. We use an MLP policy for all the architectures. \cref{fig:predprey_env} shows the environment and illustrations of a successful capture and a single-agent capture attempt to be penalized. A typical relative overgeneralization pathology could arise from imposing the single-agent capture attempt and a lack of proper coordination is that all agents crowd to a corner of the grids, failing to explore strategies. 

\Cref{fig:predprey_avgreturn} shows the average return for test episodes for varying penalties $p$ averaged over $5$ runs. Overall, DICG performs the best and solves relative overgeneralization with its implicit coordination.
Without any penalty ($p = 0$), fully centralized (CENT-MLP) and fully decentralized (DEC-MLP) architectures have similar performance. However, they require more steps to capture all prey than the DICG approaches. 
As we increase the penalty, only DICG is able to reliably and quickly converge to optimality. DEC-MLP has a characteristic slowdown in the learning curves before it is able to approach DICG. It finally converges to suboptimal performance with relative overgeneralization due to the lack of coordination across agents. The fully centralized approach CENT-MLP can only achieve positive returns in non-penalized setting. With a negative penalty, CENT-MLP cannot learn to capture prey appropriately due to two reasons: 1) simple concatenation of observations in CENT-MLP leads to a large joint observation space, and 2) concatenation of observations is not an efficient way to integrate information and learn coordination across agents. 

\noindent \textit{\textbf{Analyzing Implicit Coordination Graph}}\\
We examine the effects of the implicit CG in the following experiments.

\textbf{Semantics of implicit CG}: To understand how the DICG learns to coordinate, we perform attention weight analysis, i.e. we study the strength of soft edges of the implicit coordination. A heuristic of what affects the strength of connection between agents is the distance between agents. \Cref{fig:predprey_attnweights} shows the relationship between attention weight and distance between agents learned by DICG. Zero distance corresponds to the attention weight of an agent to itself. As the penalty increases, agents tend to increase the attention weight towards agents further away. 
This phenomenon coincides with the coordination requirements imposed by the increase of penalty that agents should pay more attention to form groups with each other to capture prey as a team than moving alone. 

\begin{figure}[t]
    \centering
    \input{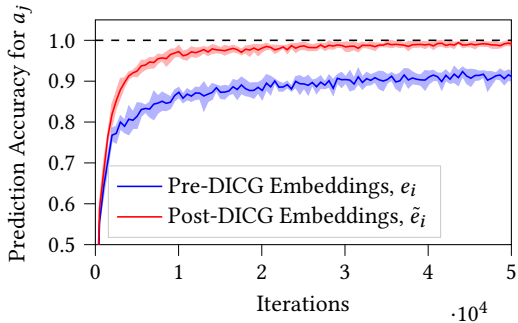}
    \caption{Predicting agent $j$'s actions ($a_j$) using agent $i$'s pre-DICG ($e_i$) and post-DICG embeddings ($\tilde{e}_i$). Data are collected from the evaluation trajectories of five random pairs of agents $(i,j)$.}
    \label{fig:action_prediction}
\end{figure}

\textbf{Effectiveness of GCN at information integration}: To examine whether information is effectively integrated across agents along the implicit coordination graph, we design an experiment to predict agent $j$'s actions $a_j$ only using agent $i$'s information with $i \neq j$. This is formulated as a supervised learning problem: $\hat{a}_j = f(x_i; \phi)$ with loss $L = CrossEntropy(\hat{a}_j, a_j)$. Theoretically, with a finite amount of data, if $x_i$ is more correlated with $a_j$, the classifier $f(\cdot; \phi)$ can produce a higher prediction accuracy. Therefore, we set $x_i$ to the pre-DICG embeddings $e_i$ and the post-DICG embeddings $\tilde{e}_i$. We use a simple MLP classifier parameterized by $\phi$, with a single 64-unit hidden layer and ReLU as activation. Results of five random pairs of $(i, j)$ are shown in \cref{fig:action_prediction}. The post-DICG embeddings can predict other agent's actions with a higher accuracy than the pre-DICG embeddings. 
We can interpret this as the DICG architecture allowing individual agents to learn other agents' intentions.

\subsection{StarCraft II Multi-agent Challenge (SMAC)}
StarCraft II, and the StarCraft II Learning Environment (SC2LE), has provided an environment for some of the most important reinforcement learning work in recent years \citep{vinyals2017starcraft, vinyals2019grandmaster}. 
However most of this work has resided in the single-agent domain where reinforcement learning is used to train a single, centralized decision-making agent how to play the entire StarCraft II game involving the control of a large number of units within the game. %
The StarCraft Multi-Agent Challenge (SMAC) extends SC2LE by providing a collection of reinforcement learning benchmarks designed specifically for \emph{multi-agent environments} \citep{samvelyan2019starcraft}. %

Within SMAC, each unit is controlled by its own separate learning agent whose actions must be conditioned on local observations and not the global game state. SMAC scenarios are designed to explore \emph{micromanagement}, e.g. precise movements and coordinated targeting, of relatively small groups of units. A scenario where 3 agent controlled stalkers combating with 5 computer controlled zealots is shown in \cref{fig:smac_env}. In each episode, positive rewards are given for the positive health point difference between the controlled agent team and the computer controlled opponent team, otherwise, the agent team receive zero or negative rewards. Large positive terminal reward is given for winning the episode by eliminating the opponents (zero for being eliminated).  

\begin{figure*}[t]
    \centering
    \input{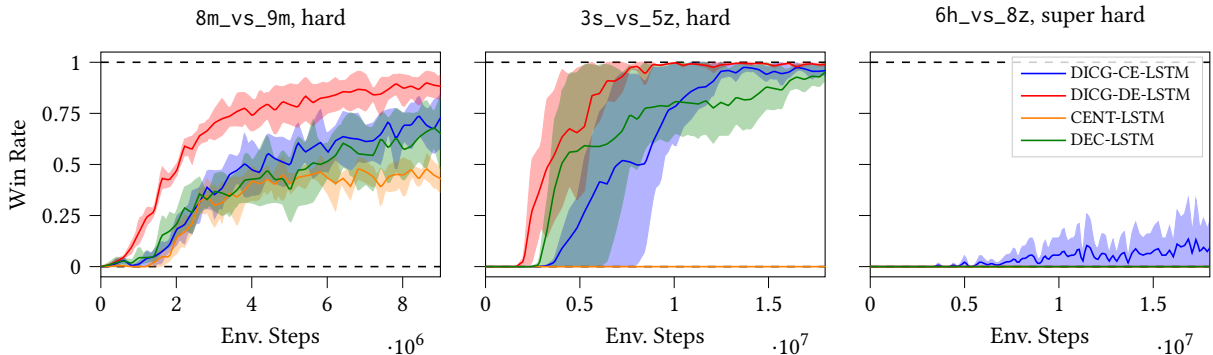}
    \caption{Evaluation win rate during training in SMAC maps.}
    \label{fig:smac}
    \vspace{-1.5em}
\end{figure*}

\begin{table}
\caption{SMAC win rate comparison.} 
\label{tab:smac}
\centering
\small
    \begin{tabular}{lccc}
        \toprule
        Approach & \texttt{8m\_vs\_9m} & \texttt{3s\_vs\_5z} & \texttt{6h\_vs\_8z}  \\ \midrule
        DCG~\citep{Bohmer2019-zv} &$55 \pm 10 \%$ &$85 \pm 3 \%$  &$\mathbf{10 \pm 5 \%}$  \\
        VDN~\citep{sunehag2018value} &$49 \pm  5\%$ &$72 \pm  10\%$  &$0$  \\
        QMIX~\citep{rashid2018qmix}  &$60 \pm  11\%$ &$95 \pm  1\%$  &$5 \pm  5\%$  \\
        CENT-LSTM                 &$42 \pm  6\%$  &0             &0  \\
        DEC-LSTM                  &$65 \pm 16\%$  &$94 \pm 5\%$  &0  \\
        DICG-CE-LSTM              &$72 \pm 11\%$  &$96 \pm 3\%$  &$9 \pm 9 \%$  \\
        DICG-DE-LSTM              &$\mathbf{87 \pm  6\%}$  &$\mathbf{99 \pm 1\%}$       &0 \\
        \bottomrule
    \end{tabular}
\end{table}
We test DICG on SMAC's asymmetric and ``micro-trick'' scenarios such as \texttt{8m\_vs\_9m}, \texttt{3s\_vs\_5z}, and \texttt{6h\_vs\_8z} \citep{samvelyan2019starcraft}. The opponent AI difficulty is set to ``hard'', ``hard'', and ``super hard'', respectively. We use an LSTM policy for all architectures in SMAC. Results of SMAC are in~\cref{fig:smac}. In \texttt{8m\_vs\_9m}, DICG-DE-LSTM outperforms all the other approaches; DICG-CE-LSTM and DEC-LSTM have similar performance; CENT-LSTM performs the worst. In \texttt{3s\_vs\_5z}, DICG-DE-LSTM shows the highest and the most stable win rate, as well as the most sample efficient learning; DICG-CE-LSTM has more stable performance than DEC-LSTM, but CENT-LSTM fails to learn. From game replays, we observe that DICG learns a particular circular movement strategy in \texttt{3s\_vs\_5z}. Due to the asymmetric setup, the 3 stalker agents controlled by DICG cannot overcome the opposing 5 zealots with force. The DICG agents learn to split into two groups, each attracting a number of opponents. Each group moves along the edges of the square map in a circle, and damages opponents using a learned hit-and-run tactic with their high speed. When the opponents are sufficiently weak, the two groups reunite to eliminate the opponents. This highly coordinated tactic demonstrates the effectiveness of DICG. In \texttt{6h\_vs\_8z}, a very difficult map, DICG-CE-LSTM is the only approach to win against the opponent AI. 

A comparison of SMAC win rate under the same difficulty setting with DCG~\citep{Bohmer2019-zv}, VDN~\citep{sunehag2018value} and QMIX~\citep{rashid2018qmix} is in~\cref{tab:smac}. 
DICG outperforms DCG and others \emph{without using the privileged state information} in \texttt{8m\_vs\_9m} and \texttt{3s\_vs\_5z}, and having comparable but noisier win rate for \texttt{6h\_vs\_8z}. DICG outperforms VDN and QMIX in all the mentioned scenarios requiring high coordination.
In many multi-agent tasks, we do not have access to privileged full state information even during training and our algorithm needs to just work with observations.
DICG demonstrates the advantage of integrated information to prevent relative overgeneralization in such scenarios.

\subsection{Traffic Junction}
The traffic junction environment, introduced by~\citet{sukhbaatar2016learning}, is a multi-agent environment where cars are randomly added to traffic junctions with pre-assigned routes who need to avoid collision with each other and reach their destinations. 
Each agent only has a limited vision of one grid from itself. 
The reward function consists of a collision penalty to discourage collision and a step cost to discourage congestion. 
There are easy, medium, and hard difficulty modes in the traffic junction environment. 
The environment configurations are adopted from~\citet{singh2018learning}. \Cref{fig:smac_env} shows the traffic junction environment in hard mode.
We use MLP policies for the traffic junction environment. 

\begin{figure*}[t!]
    \centering
    \input{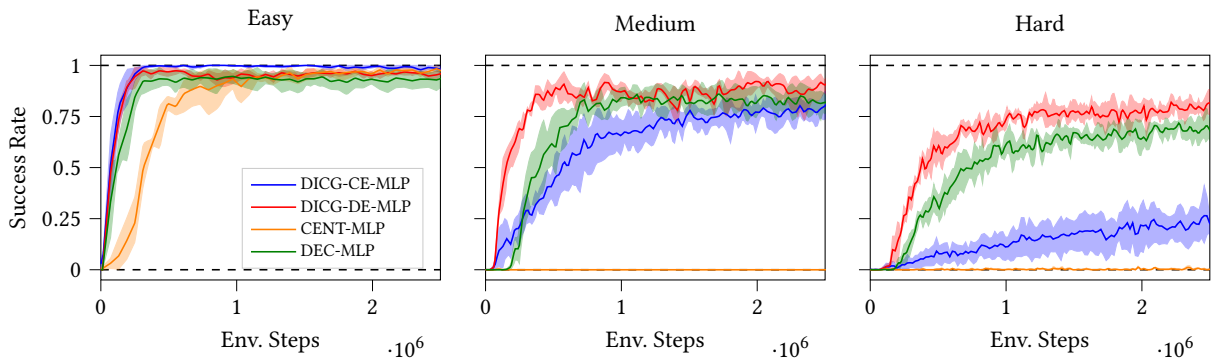}
    \caption{Evaluation success rate during training of different difficulty levels in traffic junction.}
    \label{fig:traffic}
\end{figure*}

\begin{table}
\caption{Traffic junction success rate comparison.}
\label{tab:traffic}
\centering
\small
    \begin{tabular}{lccc}
        \toprule
        Approach                  & Easy          & Medium        &Hard  \\ \midrule
        CommNet~\citep{sukhbaatar2016learning}  &$93.0 \pm 4.2\%$ &$54.3 \pm 14.2\%$ &$50.2 \pm 3.5\%$ \\
        IC3Net~\citep{singh2018learning}        &$93.0 \pm 3.7\%$ &$89.3 \pm 2.5\%$  &$ 72.4 \pm 9.6\%$ \\
        GA-Comm~\citep{liu2019multi}            &$\mathbf{99.7\%}$ &$\mathbf{97.6\%}$ &$\mathbf{82.3\%}$\\ 
        CENT-MLP                                &$97.7 \pm 0.9\%$ &0                 &0 \\
        DEC-MLP                                 &$90.2 \pm 6.5\%$ &$81.3 \pm 4.8\%$  &$69.4 \pm 4.9\%$ \\
        DICG-CE-MLP                             &$\mathbf{98.1 \pm 1.9}\%$ &$80.5 \pm 6.8\%$  &$22.8 \pm 4.6\%$ \\
        DICG-DE-MLP                             &$95.6 \pm 1.5\%$ &$90.8 \pm 2.9\%$  &$\mathbf{82.2 \pm 6.0\%}$ \\
        \bottomrule
    \end{tabular}
\end{table}
The results are in~\cref{fig:traffic} and a comparison with other baselines using the same environment configurations is in~\cref{tab:traffic}. 
DICG-CE-MLP performs better than DICG-DE-MLP in easy mode. 
CENT-MLP also performs well in easy mode. This is because easy mode has fewer number of agents and small observation space.
Centralized execution can outperform decentralized approaches in relatively small domains. 
However, DICG-CE-MLP has the privilege of more efficient agent information integration over CENT-MLP. 

In medium and hard mode, where the number of agents and the dimension of observation space increases, centralized approaches fail to perform well. 
DICG-DE-MLP outperforms decentralized and centralized baselines in medium and hard mode. 
Even though we \emph{do not use any curriculum}~\citep{bengio2009curriculum} based training, DICG's performance is close to that of GA-Comm~\citep{liu2019multi} which employs curriculum learning. 
Note that the results of GA-Comm fall within the uncertainty range of our results in easy and hard mode. We provide brief descriptions of these baselines in \cref{app:baseline_dets}.

\subsubsection*{Ablation} 
\begin{figure}[t]
    \centering
    \input{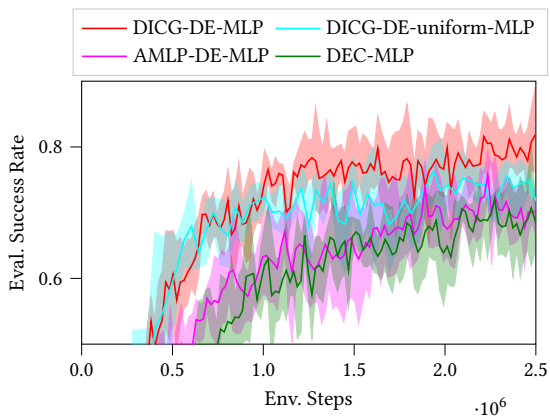}
    \caption{Zoomed-in ablation experiment results in hard mode traffic junction environment (average of 5 random seeds).}
    \label{fig:ablation_traffic}
\end{figure}

To examine the effectiveness of various components in the DICG architecture, we perform ablation experiments by adjusting the attention module and the graph convolution module. Two variants of DICG-DE-MLP are designed:
\begin{enumerate}
    \item Replacing learned attention weights with uniform attention weights, i.e. for $n$ agents, the attention weights become $1/n$. We denote this variant as DICG-DE-uniform-MLP.
    \item Replacing the graph convolution module with MLP. We concatenate the embeddings and the attention weights of all the agents and feed the concatenation through an MLP to estimate the baseline. We denote this variant as AMLP-DE-MLP.
\end{enumerate}
We test the variants in the hard mode traffic junction environment where the number of agents is large. We expect the attention module and graph convolution modules to play relatively more important roles in coordinating agents. 
The results are shown in~\cref{fig:ablation_traffic} (averaged over 5 random seeds) in a zoomed-in view. AMLP-DE-MLP shows similar performance as DEC-MLP. This indicates that MLP cannot integrate agents' information as effectively as graph convolution. DICG-DE-uniform-MLP has slightly worse performance than DICG-DE-MLP. This indicates that learned attention weights can better emphasize coordination among agents than uniformly spreading attention. \Cref{tab:ablation_traffic} compares the success rate with other approaches.

\begin{table}
\caption{Traffic junction success rate comparison with ablation experiments in hard mode.}
\label{tab:ablation_traffic}
\centering
\small
    \begin{tabular}{lc}
        \toprule
        Approach                                &Success Rate  \\ \midrule
        CommNet~\citep{sukhbaatar2016learning}  &$50.2 \pm 3.5\%$ \\
        IC3Net~\citep{singh2018learning}        &$ 72.4 \pm 9.6\%$ \\
        GA-Comm~\citep{liu2019multi}            &$\mathbf{82.3\%}$\\ 
        CENT-MLP                                &0 \\
        DEC-MLP                                 &$69.4 \pm 4.9\%$ \\
        DICG-CE-MLP                             &$22.8 \pm 4.6\%$ \\
        DICG-DE-MLP                             &$\mathbf{82.2 \pm 6.0\%}$ \\
        DICG-DE-uniform-MLP                     &$72.0 \pm 1.3\%$ \\
        AMLP-DE-MLP                             &$70.3 \pm 3.8\%$ \\
        \bottomrule
    \end{tabular}
\end{table}

\section{Conclusions and Future Work}
In this work, we present the DICG architecture that uses self-attention to implicitly build a coordination graph and then perform message passing with graph convolution layers to compute appropriate baseline values (DE) or actions (CE) for the agents while keeping the computational graph differentiable.
We demonstrate that DICG solves the relative overgeneralization pathology in predator-prey tasks, as well as various MARL baselines including the challenging StarCraft II micromanagement tasks and traffic junction tasks. 
DICG is shown to be an effective architecture for implicitly and dynamically learning multi-agent coordination that achieves an appropriate tradeoff between fully centralized and fully decentralized approaches.
For future work, we aim to improve the sample efficiency of DICG. To achieve this, we may incorporate the DICG architecture into off-policy learning algorithms such as deep Q-learning~\citep{mnih2013playing} and soft actor-critic~\citep{haarnoja2018soft,wei2018multiagent}. 


\begin{acks}

This material is based upon work supported by the Under Secretary of Defense for Research and Engineering under Air Force Contract No. FA8702-15-D-0001. Any opinions, findings, conclusions or recommendations expressed in this material are those of the author(s) and do not necessarily reflect the views of the Under Secretary of Defense for Research and Engineering.


\end{acks}



\bibliographystyle{ACM-Reference-Format} 
\bibliography{main}


\clearpage
\appendix

\section{Experiment Details}\label{app:exp_details}
\begin{table}[h]
\caption{PPO~\citep{schulman2017proximal} and optimizer parameters.}
\label{tab:ppo_params}
\centering
\small
    \resizebox{0.9\columnwidth}{!}{\begin{tabular}{cc}
        \toprule
        Parameter  &Value\\ \midrule
        Likelihood ratio clip range &$0.2$ \\
        Optimizer &Adam~\citep{kingma2014adam} (PyTorch~\citep{pytorch})\\
        Policy learning rate &$3 \times 10^{-4}$\\
        GAE-$\lambda$ &$0.97$\\
        Discount $\gamma$ &$0.99$\\
        Trajectory length &Environment dependent\\ 
        Batch size &Environment dependent\\ 
        Policy entropy coefficient &Environment dependent\\ 
        \bottomrule
    \end{tabular}}
\end{table}

This section describes the environment settings, hyperparameters of the policy optimization algorithm, and network sizes. \cref{tab:ppo_params} shows the hyperparameters used for PPO shared by all experiments. Batch size is in the unit of environment steps. All the nonlinear activation functions in this work are hyperbolic tangent ($\tanh$). The aggregator network used for DICG-DE approach is a single layer linear mapping from the embedding dimension to one dimension. Environment specific detailed settings are listed in following subsections.

Code of this work is available at this link\footnote{Code of this work: \url{https://github.com/sisl/DICG}}.

\subsection{Predator-Prey}
We use the implementation of predator-prey from open source \texttt{ma-gym}.\footnote{\url{https://github.com/koulanurag/ma-gym/tree/master/ma_gym/envs/predator_prey}} 
The original implementation does not include other predators in the observation of a predator. We modify the environment by making the other agents visible in the field of view of a predator. A predator has a field of view of $5 \times 5$ grids with itself at center. 

\begin{table*}
\caption{Predator-prey network architectures.}
\label{tab:predprey_params}
\centering
\small
    \begin{tabular}{ccccc}
        \toprule
        Approach     &DICG encoder sizes &DICG embedding size &MLP policy sizes &Baseline (critic) sizes\\ \midrule
        DICG-CE-MLP  &$[128]$       &$64$               &$[128, 64, 32]$        &$[64, 64, 32]$\\
        DICG-DE-MLP  &$[128]$       &$64$               &$[128, 64, 32]$        &$[64, 64, 32]$\\
        DEC-MLP      &N/A           &N/A                &$[128, 64, 32]$        &$[64, 64, 32]$\\
        CENT-MLP     &N/A           &N/A                &$[512, 128, 64]$       &$[64, 64, 32]$\\
        \bottomrule
    \end{tabular}
\end{table*}
We train with network settings shown in~\cref{tab:predprey_params}. The maximum number of steps in an episode is set to 200. We use a batch size of $6\times 10^4$ and a policy entropy coefficient of $0.1$. Except for DICG-DE-MLP, all other approaches use an MLP baseline (critic). Two layers of GCN are used for DICG.

\subsection{StarCraft II Multi-agent Challenge (SMAC)}
We use the open source SMAC implementation by~\citet{samvelyan2019starcraft}.
We use the SMAC implementation\footnote{\url{https://github.com/oxwhirl/smac}} by~\citet{samvelyan2019starcraft}. 
The number of controlled agents and opponents, their unit types as well as maximum time steps of an episode are redefined by the authors of the micromanagement scenarios. The detailed map configurations and the PPO parameters we use are listed in~\cref{tab:smac_maps}. For the network architecture, two layers of GCN are used for DICG. Other network architecture details are listed in~\cref{tab:smac_params}. 
Except for DICG-DE-LSTM, all other approaches use an MLP feature baseline (critic).

\begin{table*}
\caption{SMAC scenario (map) information and training settings.}
\label{tab:smac_maps}
\centering
\small
    \begin{threeparttable}
    \begin{tabular}{ccccccc}
        \toprule
        Map &Difficulty &Controlled agents &Opponents &Max steps &Batch size &Policy entropy coeff.\\ \midrule
        \texttt{8m\_vs\_9m} &Hard &8 marines  &9 marines  &$120$ &$8\times10^4$ &$0.01$\\
        \texttt{3s\_vs\_5z} &Hard &3 stalkers &5 zealots  &$250$ &$6\times10^4$ &$0.025/0.1$\tnote{*}\\
        \texttt{6h\_vs\_8z} &Super hard &6 hydralisks  &8 zealots  &$150$ &$6\times10^4$ &$0.025/0.1$\tnote{*}\\
        \bottomrule
    \end{tabular}
    \begin{tablenotes}
      \small
      \item * For DICG-DE-LSTM.
    \end{tablenotes}
    \end{threeparttable}
\end{table*}

\begin{table*}
\caption{SMAC network architectures.}
\label{tab:smac_params}
\centering
\small
    \begin{tabular}{cccccc}
        \toprule
        Approach      &DICG encoder sizes &DICG embedding size &Policy encoder sizes &LSTM hidden size &Baseline (critic) sizes\\ \midrule
        DICG-CE-LSTM  &$[128]$  &$128$ &$[128]$ &$64$ &$[64, 64, 64]$\\
        DICG-DE-LSTM  &$[128]$  &$64$  &$[128]$ &$64$ &$[64, 64, 64]$\\
        DEC-LSTM      &N/A      &N/A   &$[128]$ &$64$ &$[64, 64, 64]$\\
        CENT-LSTM     &N/A      &N/A   &$[256]$ &$128$ &$[64, 64, 64]$\\
        \bottomrule
    \end{tabular}
\end{table*}

\subsection{Traffic Junction}
We use the open source traffic junction environment implementation by \citet{sukhbaatar2016learning}. 
We use the maximum agent add rate (the most difficult task setting) from their curriculum framework and skip curriculum training.
We use a policy entropy coefficient of $0.02$. Except for DICG-DE-MLP, all other approaches use an MLP baseline (critic). The detailed environment settings are listed in~\cref{tab:traffic_config}. The vision range of cars (agents) is set to one grid. The network configurations we used are in~\cref{tab:traffic_params}.

\begin{table*}
\caption{Traffic junction environment configurations~\citep{singh2018learning}.}
\label{tab:traffic_config}
\centering
\small
    \begin{tabular}{ccccccccc}
        \toprule
        Difficulty  &\# roads &\# directions & Road dim. &\# junctions &$n_\text{max}$ &Car add rate &Max steps &Batch size\\ \midrule
        Easy    &$2$ &$1$   &$7$ &$1$ &$5$    &$0.3$    &$20$ &$6\times10^4$\\
        Medium  &$4$   &$2$  &$14$ &$1$   &$10$    &$0.2$   &$40$ &$6\times10^4$\\
        Hard    &$8$  &$2$  &$18$  &$4$  &$20$    &$0.05$   &$60$ &$8\times10^4$\\
        \bottomrule
    \end{tabular}
\end{table*}

\begin{table*}
\caption{Traffic junction network architectures.}
\label{tab:traffic_params}
\centering
\small
    \begin{tabular}{cccccc}
        \toprule
        Approach     &DICG encoder sizes &DICG embedding size &MLP policy sizes &Baseline (critic) sizes\\ \midrule
        DICG-CE-MLP  &$[128]$       &$128$      &$[128, 64, 32]$  &$[64,64,64]$\\
        DICG-DE-MLP  &$[128,128]$   &$128$      &$[256, 128, 64]$ &N/A\\
        DEC-MLP      &N/A           &N/A        &$[256, 128, 64]$ &$[64,64,64]$\\
        CENT-MLP     &N/A           &N/A        &$[512, 128, 64]$ &$[64,64,64]$\\
        \bottomrule
    \end{tabular}
\end{table*}

\section{Descriptions for Compared Baseline Approaches}
\label{app:baseline_dets}
We compared with the following baseline approaches:
\begin{itemize}
    \item Value-decomposition Networks (VDN)~\citep{sunehag2018value}: Value-based off-policy method that factorizes the joint value function as a sum of individual agent value functions.
    \item QMIX~\citep{rashid2018qmix}: Value-based off-policy method that factorizes the joint value function as a monotonic function of the individual agent-value functions.
    \item Deep Coordination Graphs (DCG)~\citep{Bohmer2019-zv}: Value-based off-policy method to learn factored value representations based on pre-determined static coordination graph.
    \item CommNet~\citep{sukhbaatar2016learning}: A multi-agent communication model uses multi-step centralized and aggregated (sum and mean) communication channels to share information among agents. This model increases the receptive field of agents through communication. An on-policy method.
    \item IC3Net~\citep{singh2018learning}: An extension of CommNet with more sophisticated communication model. An on-policy method.
    \item GA-Comm~\citep{liu2019multi}: An on-policy communication modeling method. It first uses hard attention to select which agents to communicate with, then uses soft attention and GCN to get contribution of other agents.

\end{itemize}

\end{document}